\documentclass{article}

\usepackage{microtype}
\usepackage{graphicx}
\usepackage{subfigure}
\usepackage{booktabs} 

\usepackage{hyperref}

\def \adapterlogo {\raisebox{-0.1\height}{\includegraphics[height=0.95\baselineskip]{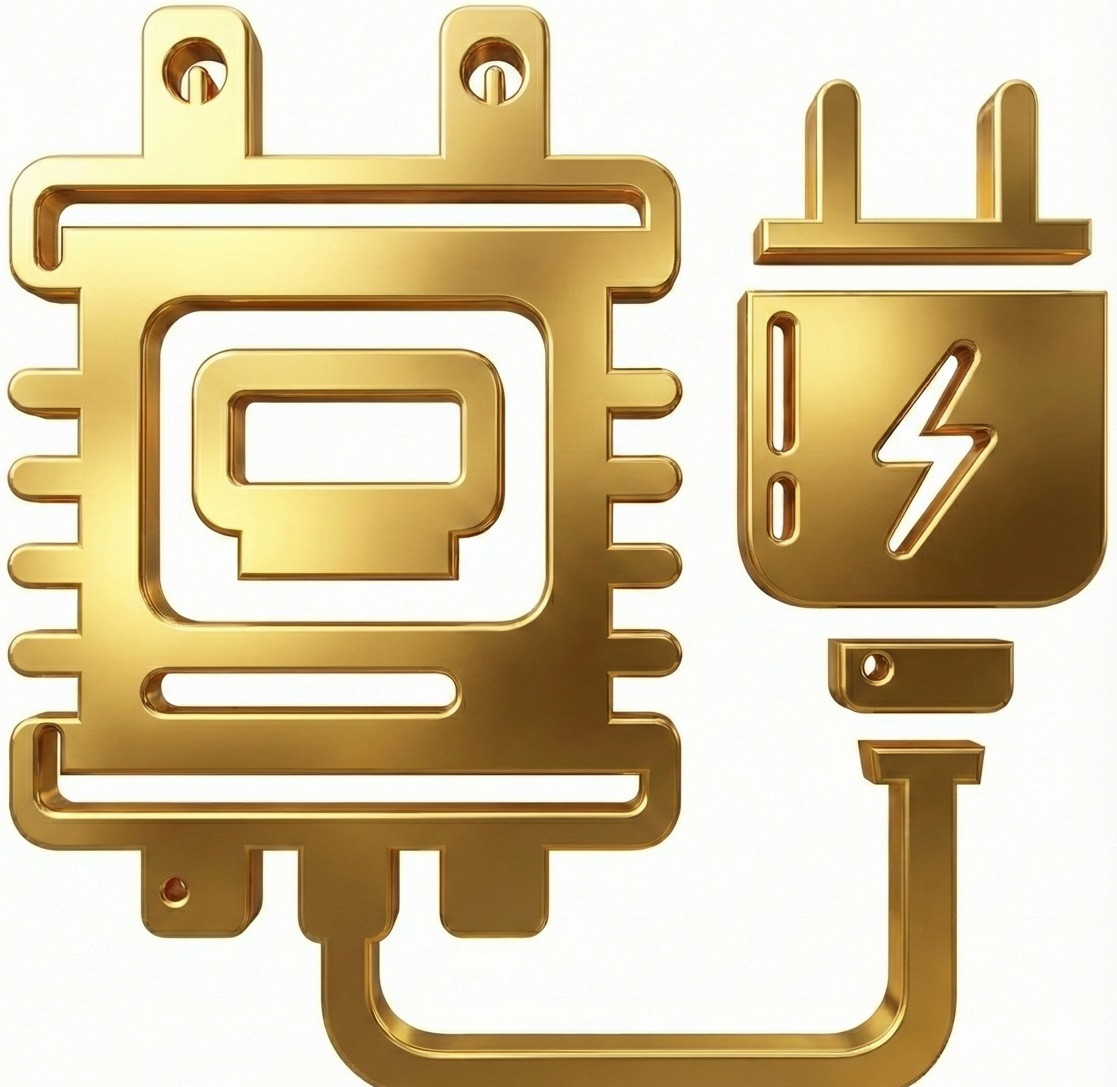}}}

\usepackage[preprint]{icml2026}

\usepackage{amsmath}
\usepackage{amssymb}
\usepackage{mathtools}
\usepackage{amsthm}

\usepackage[capitalize,noabbrev]{cleveref}

\theoremstyle{plain}

\theoremstyle{definition}

\theoremstyle{remark}

\usepackage[textsize=tiny]{todonotes}

\usepackage[table]{xcolor}
\usepackage{multirow}
\usepackage{amsmath}
\usepackage{bm}

\icmltitlerunning{1\%\textgreater 100\% : High-Efficiency Visual Adapter with Complex Linear Projection Optimization}

\begin{document}

\twocolumn[
\icmltitle{\adapterlogo{} 1\%\textgreater 100\% : High-Efficiency Visual Adapter with \\Complex Linear Projection Optimization
}





\begin{icmlauthorlist}
\icmlauthor{Dongshuo Yin}{thu}
\icmlauthor{Xue Yang}{sjtu}
\icmlauthor{Deng-Ping Fan}{nku}
\icmlauthor{Shi-Min Hu}{thu}
\end{icmlauthorlist}

\icmlaffiliation{thu}{BNRist, Department of Computer Science and Technology, Tsinghua University, Beijing, China}
\icmlaffiliation{sjtu}{School of Automation and Intelligent Sensing, Shanghai Jiao Tong University, Shanghai, China}
\icmlaffiliation{nku}{Nankai lnternational Advanced Research Institute (Shenzhen Futian) \& SLAI, Shenzhen, China}


\icmlkeywords{Machine Learning, ICML}

\vskip 0.3in
]



\printAffiliationsAndNotice{} 

\begin{abstract}

Deploying vision foundation models typically relies on efficient adaptation strategies, whereas conventional full fine-tuning suffers from prohibitive costs and low efficiency. While delta-tuning has proven effective in boosting the performance and efficiency of LLMs during adaptation, its advantages cannot be directly transferred to the fine-tuning pipeline of vision foundation models. To push the boundaries of adaptation efficiency for vision tasks, we propose an adapter with \textbf{Co}mplex \textbf{Lin}ear Projection Optimization \textbf{(CoLin)}. For architecture, we design a novel low-rank complex adapter that introduces only about 1\% parameters to the backbone. For efficiency, we theoretically prove that low-rank composite matrices suffer from severe convergence issues during training, and address this challenge with a tailored loss. Extensive experiments on object detection, segmentation, image classification, and rotated object detection (remote sensing scenario) demonstrate that CoLin outperforms both full fine-tuning and classical delta-tuning approaches with merely 1\% parameters for the first time, providing a novel and efficient solution for deployment of vision foundation models. We release the code on \href{https://github.com/DongshuoYin/CoLin}{https://github.com/DongshuoYin/CoLin}.


\end{abstract}


\begin{figure}
	\centering
	\includegraphics[scale=.26]{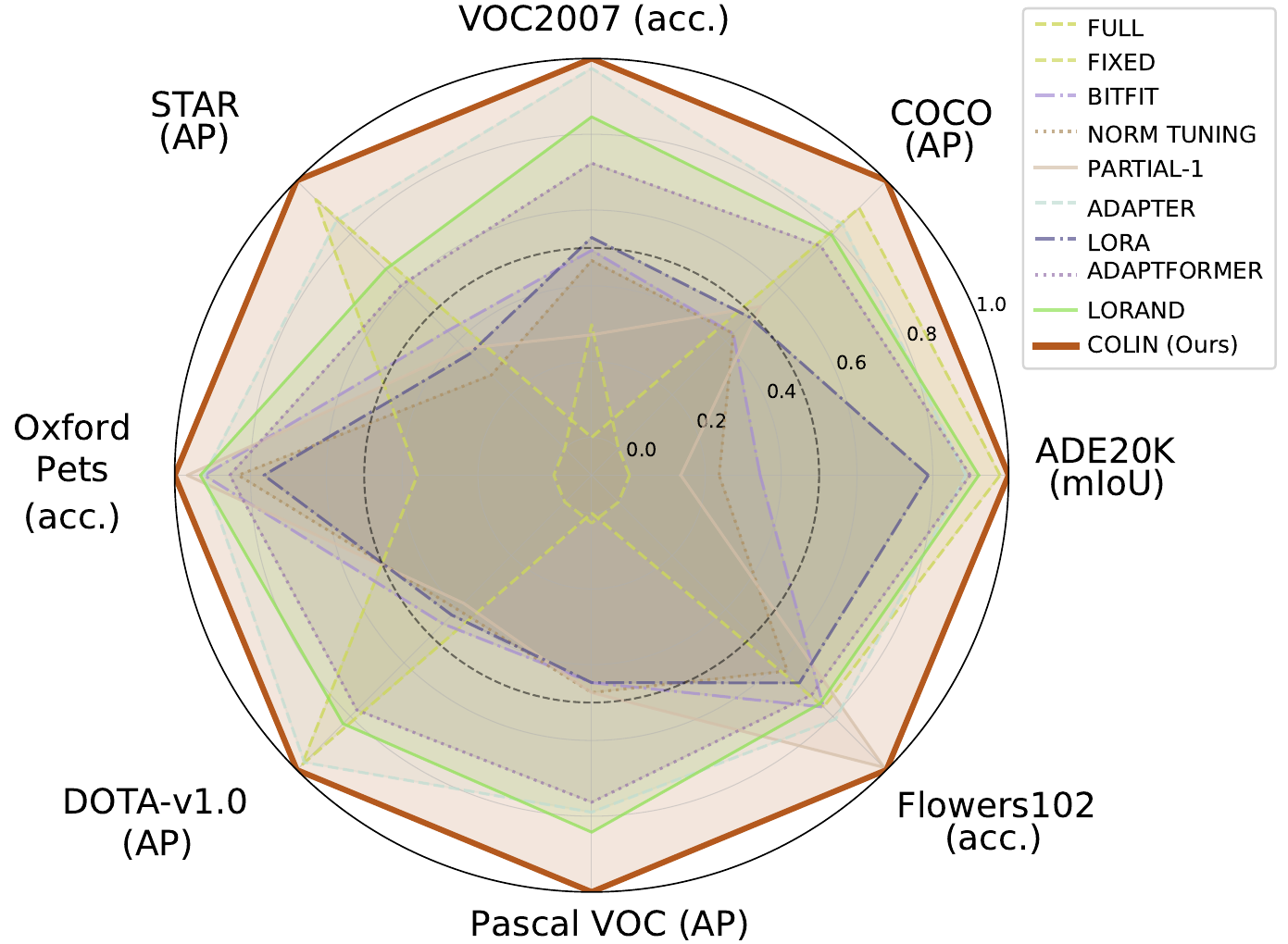}
	\caption{\textbf{Performance radar.} The proposed method outperforms full fine-tuning and typical delta-tuning approaches (based on Swin-B/L) on 8 visual tasks by fine-tuning \textbf{only about 1\%} new params for the first time. This parameter efficiency advantage can be further enhanced on larger foundation models. The maximum on each axis represents the best performance of each dataset.
}
	\label{fig:radar}
\end{figure}

\section{Introduction}

Explosive growth of visual \cite{awais2025foundation} and multimodal \cite{li2025benchmark} foundation models is spearheading an AI paradigm shift. Endowed with universal visual representation capabilities from large-scale pre-training \cite{du2022survey}, these models demonstrate strong generalization on core tasks (e.g., segmentation, detection) and empower high-stakes domains including autonomous driving \cite{gao2024survey}, remote sensing \cite{yin2025remote, hu2024airs, hu2024tea}, and medical imaging \cite{dutt2023parameter}. The distribution shift between pre-trained tasks and downstream tasks makes model adaptation a linchpin for real-world deployment. While full fine-tuning aligns task performance, it incurs prohibitive computational and storage costs and risks catastrophic forgetting. Delta-tuning \cite{han2024parameter, yu2024visual} enables lightweight adaptation by freezing backbones and optimizing only a small parameter subset—has emerged as a research focus. Efficient visual model adaptation is thus pivotal to foundation model deployment. In visual tasks, how to outperform full fine-tuning while updating the minimal number of parameters is a research question of significant theoretical and practical importance.

Recent delta-tuning methods in computer vision can be broadly divided into three families: reparameterization, adapter-based approaches, and partial fine-tuning. Reparameterization methods, such as the LoRA series \cite{hu2021lora}, have shown impressive gains in large language models, but these benefits do not easily carry over to visual foundation models. Partial-tuning only updates certain components of the foundation model (such as normalization layers \cite{giannou2023expressive}, biases \cite{zaken2021bitfit}, or the final few transformer blocks \cite{yosinski2014transferable}), which often falls short of full fine-tuning in terms of performance \cite{yin2023parameter}. Adapter-based methods \cite{houlsby2019parameter}, while adding some inference overhead, generally yield more competitive results compared to the other two strategies. Original adapters consist of two linear layers: an up-projection and a down-projection. Together, these layers account for most of the adapter’s parameter, and their transfer efficiency is crucial for downstream task performance \cite{yin20231}. However, existing art has not systematically analyzed or optimized linear projection layers for visual recognition tasks, limiting the efficiency ceiling of adapters in both vision tasks and even multimodal tasks.

To further improve the efficiency and performance of adapters for visual tasks, we conduct an in-depth analysis and optimization of the linear layers within the adapter architecture. First, we propose a multi-branch low-rank adapter structure based on a sophisticated sharing mechanism. Inspired by Mixture-of-Experts (MoE) designs, we replace the original monolithic projection matrix with a sum of multiple low-rank matrices of the same size. This design enhances the robustness of the projection matrix while significantly reducing the number of newly introduced parameters. Building on this, we introduce branch sharing and kernel matrix sharing mechanisms to further decrease the parameter footprint and improve the consistency of the adapter’s visual feature understanding. Compared to full fine-tuning, our approach reduces the number of parameters by more than 98\%. Second, we observe that this design can lead to significant inefficiency during training. To investigate the factors affecting efficiency, we derive and prove, using matrix-theoretic arguments \cite{franklin2000matrix}, that the low-rank composite matrices suffer from gradient direction entanglement during backpropagation. Based on this theoretical insight, we introduce a necessary orthogonal loss during training and optimize the parameter initialization strategy to mitigate the impact of inefficient transfer. We perform extensive experiments to validate the effectiveness of our proposed method. Results (Figure \ref{fig:radar}) on object detection, instance segmentation, semantic segmentation, image classification, and rotated object detection tasks demonstrate that CoLin outperforms full fine-tuning and classical delta-tuning methods on both natural image and remote sensing scenarios. The contributions can be summarized as follows:
\begin{itemize}
    \item We propose a multi-branch low-rank adapter architecture built on a sophisticated sharing mechanism, which significantly reduces the parameter and improves the transfer efficiency of visual features.
    \item Using matrix-theoretic principles, we prove the existence of an inefficiency issue in low-rank composite matrices during gradient backpropagation and introduce an orthogonal loss to mitigate the adverse effects of inefficient gradient transfer.
    \item Comprehensive results on object detection, instance segmentation, semantic segmentation, image classification and rotated object detection demonstrate that the proposed method consistently outperforms both full fine-tuning and classical delta-tuning approaches with only about 1\% new params across all these scenarios.
        
\end{itemize}

\section{Related Work}
\subsection{Delta-tuning}

Delta-tuning has been widely adopted across large language models, vision foundation models, and multimodal large models due to its parameter-efficiency and impressive performance. In natural language processing, LoRA \cite{hu2021lora} achieves strong performance with low-rank matrices. Adapter \cite{houlsby2019parameter} insert bottleneck structures into the backbone to facilitate domain transfer, while prompt tuning optimizes model outputs by modifying the input distribution. In the vision domain, visual prompt tuning \cite{jia2022visual} treats additional learnable patches as prompts for fine-tuning. AdaptFormer \cite{chen2022adaptformer} adds scalable bottleneck structures in skip connections to improve adaptation, and DAPE \cite{xia2025dape} employs a two-stage fine-tuning strategy for video editing. In multimodal tasks, DoRA \cite{liu2024dora} extends LoRA by optimizing both magnitude and direction, leading to better performance on text and video tasks. LoRASculpt \cite{liang2025lorasculpt} mitigates catastrophic forgetting in multimodal fine-tuning through LoRA pruning. Overall, delta-tuning not only significantly reduces the number of trainable parameters but also amplifies the advantages of pre-trained models in terms of performance, continual learning, and domain adaptation.

\begin{figure*}
	\centering
	\includegraphics[scale=.8]{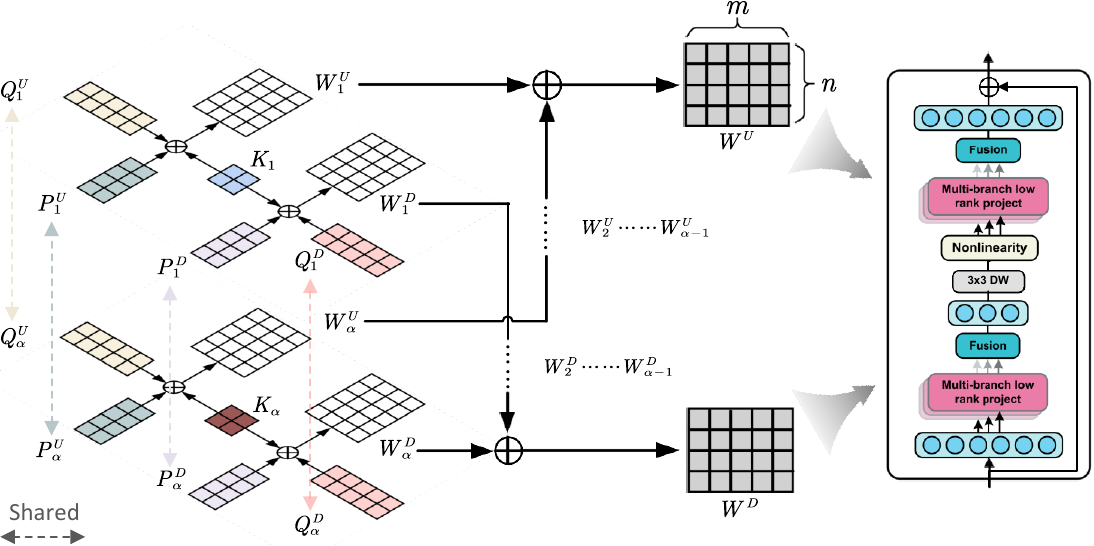}
	\caption{\textbf{Module schematic.} The down-projection $W^D$ and up-projection $W^U$ matrices are the summation of $\alpha$ branches $W_1^D(W_1^U)...W_{\alpha}^D(W_{\alpha}^U)$. $K_i$ in $i$-th branch is shared between $W_i^D$ and $W_i^U$. All $P$ and $Q$ are shared among branches. All $K_i$ are trainable, and all the $W$ matrices are calculated. A single depth-wise (DW) convolution layer is added before GeLU. }
	\label{fig:detail}
\end{figure*}

\subsection{Adapter Optimization}
Adapter-based methods have numerous variants, several of which have demonstrated impressive results in visual tasks. Appendix \ref{appen-paradigm} presents the mathematical process of the adapter-tuning. Polyhistor \cite{liu2022polyhistor} introduces a novel insertion strategy for multi-task learning. AdapterDrop \cite{ruckle2021adapterdrop} dynamically removes lower-layer adapters during training to reduce computational overhead. LoRand \cite{yin20231} is the first work to apply adapter tuning to dense prediction tasks in vision. E$^3$VA \cite{yin2023parameter} proposes an adapter highway for visual tasks, effectively lowering training costs. ST-Adapter \cite{pan2022st} extends adapter-based adaptation to image-to-video tasks by incorporating depthwise 3D convolutions. Compacter \cite{karimi2021compacter} leverages Kronecker products to reduce the parameter count of adapters. The linear layers in adapters constitute the primary source of their parameter overhead. To date, existing work has not yet provided theoretical analysis or dedicated optimization for the linear layers of visual adapters in vision tasks.

\section{Methods}



\subsection{Multi-branch Low-rank Projection}

\subsubsection{Standard Adapter Linear Projection}
Before introducing the proposed method, we first review the existing adapter structure. Conventional adapters are bottleneck structures containing a down-projection, an up-projection, and a non-linear activation function. Besides, adapters ensure the robustness of the model by adding residual \cite{he2016deep} structures. Adapter layer can be formulated as follows:
\begin{equation}
A^l=U^l\left(GeLU\left(D^l\left(x\right)\right)\right)+x,
\end{equation}
where $U^l$ and $D^l$ represent the up and down projections in the $l$-th adapter layer, and GeLU is the activation function. It is clear that the parameters in adapter come from the projections. The projection process can be written as:
\begin{equation}
y=Wx+b,
\end{equation}
which means most adapter parameters are in $W$.

\subsubsection{Low-rank Linear Projection}
\label{sec:LLP}

To reduce the adapter parameters, we propose a projection matrix construction method based on multi-branch design, which can effectively reduce the parameters of $W$. Figure \ref{fig:detail} shows the simplified structure of CoLin. Here we approximate not a specific matrix $W$ but an ideal matrix $W_{best}$ that can transform the feature space of the pre-trained model into new tasks by heuristic learning. The approximation matrix $\hat{W}$ has the same size as $W$, but the low-rank design makes $\hat{W}$ have far fewer free degrees than a common $W$.

Specifically, we synthesize each $W$ by multiplying three low-rank matrices $P\in\mathbb{R}^{\beta\times m},K\in\mathbb{R}^{\beta\times\beta},\ Q\in\mathbb{R}^{\beta\times n}$, that is:
\begin{equation}
W=P^TKQ,
\end{equation}
where $ \beta\ll\min(m,n)$ ensuring that $P$ and $Q$ are low-rank matrices. $K$ can be regarded as a kernel matrix that controls the parameter size of CoLin.

\subsubsection{Multi-branch Linear Projection}


The low-rank design offers two primary advantages. \textbf{1) Significant parameter reduction}: For $n$-dimensional visual features, the projection matrix of a vanilla adapter (with intermediate dimension as half the input dimension) requires \(n^2/2\) parameters, whereas our projection matrix in Section \ref{sec:LLP} only needs \((n + n/2)/\beta = 3n\beta/2\) parameters. Taking \(n=768\) (common in vision models) and \(\beta=8\), the low-rank matrix achieves 97\% parameter reduction. \textbf{2) Mitigating parameter redundancy}: As revealed in previous works \cite{hu2021lora, karimi2021compacter}, the feature space of superior pre-trained models substantially exceeds the requirements of most downstream tasks. This necessitates adaptation in low-dimensional subspaces to identify optimal parameter configurations. Vanilla adapters exhibit unnecessary parameter freedom, leading to inefficient adaptation and potential overfitting in few-shot scenarios. Low-rank matrices inherently constrain this redundancy.

However, extremely low-parameter matrices may confine adapters to suboptimal subspaces, limiting overall performance. To address this without increasing inference latency, we introduce a multi-branch structure inspired by Mixture-of-Experts (MoE) \cite{riquelme2021scaling} and AdaBoost \cite{sagi2018ensemble, thilagavathi2021evaluating} principles. Each branch \(\{P_i^TK_iQ_i\}\) captures distinct feature interpretations at its feature level, with collaborative branching enhancing adaptation robustness and decision stability. To maintain inference efficiency, we implement branch fusion through parameter summation:

\begin{equation}
W = \sum_{i=1}^{\alpha}W_i = \sum_{i=1}^{\alpha}{P_i^TK_iQ_i}.
\end{equation}

This architecture enables: 1) Expanded solution space through branch diversity, 2) Error compensation between branches via ensemble effects, and 3) Preserved computational efficiency by linear combination. We insert CoLin into each SwinBlock twice (see Appendix \ref{pos} for details).

\subsection{Complex Sharing Strategy}

Compacter \cite{karimi2021compacter} demonstrates that appropriate sharing mechanisms can enhance the efficiency and performance of adapters. To further optimize the parameter efficiency and robustness of low-rank adapters, we design two sharing mechanisms for multi-branch low-rank structures: kernel sharing and branch sharing.

\subsubsection{Kernel Sharing}

In adapters, the down-projection and up-projection layers correspond to feature compression and restoration, respectively. However, such compression may lead to loss of critical feature information. To preserve essential information during the feature dimension restoration process, we share the kernel matrix $K$ of the two projection layers within each branch. We hope the sharing mechanism can promote the coherence of two projection layers during training process. Besides, the shared $K$ also slightly reduces the number of parameters. With kernel sharing, the $W^U$ and $W^D$ in the proposed structure can be represented as:
\begin{equation}
W^U=\sum_{i=1}^{\alpha}W_i^U=\sum_{i=1}^{\alpha}{{{(P}_i^U)}^TK_i}Q_i^U,
\end{equation}
\begin{equation}
W^D=\sum_{i=1}^{\alpha}W_i^D=\sum_{i=1}^{\alpha}{{{(P}_i^D)}^TK_i}Q_i^D,
\end{equation}
where $K_i$ is shared in $W^U$ and $W^D$.

\subsubsection{Branch Sharing}
Kernel sharing is performed between the up/down projections within the same branch. For different branches, we design two distinct types of matrices. We aim for all kernel matrices to focus on branch-specific adaptations while preserving maximum flexibility. As for the higher-parameter matrices $P$ and $Q$, we intend them to capture cross-task universal features or transformations, thereby enhancing the model's generalization capability.

Finally, the $W^U$ and $W^D$ in CoLin can be represented as:
\begin{equation}
W^U=\sum_{i=1}^{\alpha}W_i^U=\sum_{i=1}^{\alpha}{{{(P}^U)}^TK_i}Q^U,
\end{equation}
\begin{equation}
W^D=\sum_{i=1}^{\alpha}W_i^D=\sum_{i=1}^{\alpha}{{{(P}^D)}^TK_i}Q^D,
\end{equation}
where $P^U$ and $Q^U$ is shared in $\alpha$ branches. Figure \ref{fig:detail} presents the detailed designs of the multi-branch projection.

\subsection{Orthogonal Optimization of the Parameter Space}
\label{PSQ}

The idea of low-rank synthesis can effectively reduce parameter redundancy, but it also makes the model's convergence process inefficient. To illustrate this, we first model the matrix $W$ as a simple form $W = PQ$. Ideally, the change of $W$ before and after an update can be expressed as follows: 
\begin{equation}
\Delta W = W' - W = -\eta \nabla_W L,
\end{equation}
where $\eta$ denotes the learning rate and $\nabla_W L$ is the gradient of the loss $L$ with respect to $W$. In practice, when $W = PQ$, the parameters actually updated are $P$ and $Q$. Thus, $\Delta W$ can be approximated by the following equation (neglecting the trivial $\eta^2$ term):

\begin{equation}
\begin{aligned}
\Delta W &= P'Q' - PQ = (P - \eta \nabla_P L)(Q - \eta \nabla_Q L) - PQ \\
&= -\eta \left( \nabla_P L \cdot Q + P \cdot \nabla_Q L \right) + \eta^2 \cdot \nabla_P L \cdot \nabla_Q L \\
&\approx -\eta \left( \nabla_P L \cdot Q + P \cdot \nabla_Q L \right).
\end{aligned}
\end{equation}

To further analyze the above equation, we need to simplify the matrix gradients using the trace trick. First, we compute $\nabla_Q L$. According to the definition of loss differential, we have:
\begin{equation}
dL = \text{tr}(\nabla_W^T dW) = \text{tr}(\nabla_W^T \cdot P \cdot dQ), 
\end{equation}
where $\text{tr}(\cdot)$ denotes the matrix trace. Using the cyclic permutation property of the trace $\text{tr}(AB) = \text{tr}(BA)$, the differential can be rewritten as $dL = \text{tr}\left( (P^T \nabla_W)^T \cdot dQ \right)$. By the definition of matrix gradient, the derivative of $dL$ with respect to $dQ$ equals the transpose of the coefficient of $dQ$ in the trace. Thus, the gradient can be expressed as: 
\begin{equation}
\nabla_Q L = P^T \nabla_W.
\end{equation}
Similarly, $\nabla_P L = \nabla_W Q^T$. Substituting these gradients into the equation for $\Delta W$ gives:

\begin{equation}
\Delta W \approx -\eta \nabla_W \left( Q Q^T + P^T P \right).
\end{equation}

According to the loss differential $dL = \text{tr}(\nabla_W^T \Delta W)$, $dL$ becomes smaller when $\Delta W$ is in the same direction as $\nabla_W$. In other words, the optimization efficiency of $W$ is maximized when $P$ and $Q$ are orthogonal. Therefore, we expect $P$ and $Q$ to tend to be orthogonal, i.e., satisfying:

\begin{equation}
Q Q^T \approx I_k \quad \text{and} \quad P^T P \approx I_k.
\end{equation}
In this case, $\Delta W \approx -2\eta \nabla_W$. For detailed proof, please refer to Appendix \ref{proof}. To maintain the orthogonality of $P$ and $Q$, we add an additional loss for each $P_i$ and $Q_i$:

\begin{equation}
L_{PQ}^i = ||P^T P - I||_F^2  + ||Q Q^T - I||_F^2,
\end{equation}
where $||\cdot||_F$ denotes the Frobenius Norm. Let the original loss be $L_0$ and the proposed loss be $L_{ort}$, then the new loss with an extra hyper-parameter coefficient $\lambda$ is then:

\begin{equation}
L = L_0 + L_{ort} = L_0 + \lambda\cdot\sum_{i=1}^n L_{PQ}^i.
\end{equation}

\subsection{SVD-based Initialization}
\label{SVD}

From section \ref{PSQ}, we know that orthogonality facilitates the convergence of adapters. To enable matrices $P$ and $Q$ to better maintain orthogonality, we adopt orthogonal initialization instead of the classical random initialization. Specifically, for $P\in\mathbb{R}^{\beta\times m}$, $K\in\mathbb{R}^{\beta\times\beta}$ and $Q\in\mathbb{R}^{\beta\times n}$, we first randomly initialize a matrix $W_0\in\mathbb{R}^{m\times n}$, and then decompose $W_0$ into $U$, $S$ and $V$ via singular value decomposition (SVD), i.e., 
\begin{equation}
U,S,V = \text{SVD}(W_0). 
\end{equation}
For down-projection, we set $P_0 = S$, $K_0 = S$ and $Q_0 = V$. For up-projection, we only set $P_0 = S$ and $Q_0 = V$. We use kaiming uniform to initialize $W_0$.

\subsection{Parameter Analysis}

Parameter efficiency is a metric for evaluating adapter designs. Here, we compare the number of new parameters between the standard linear layer and the proposed linear layer in the adapter. Assume that there are $\gamma$ linear layers in the adapter, new parameters of the standard linear layer is $\gamma mn$ (ignoring the bias), while the proposed linear layer only requires $\gamma\beta(m+n)+\alpha\beta^2$, which will significantly reduce the number of new parameters in the adapter.

\begin{table*}[tb]

	\centering
    \caption{\textbf{Results of baselines and our method on Pascal VOC and ADE20K benchmarks.} Swin-L is employed as the pre-trained model here. We present the numbers and percentages of trainable backbone parameters on the left and all the performences on the right. $\ast$ denotes the trainable parameters in backbones. The best AP/mIoU in each column is bolded.}
	\scalebox{.8}{
		\setlength{\tabcolsep}{4.5mm}{
			\begin{tabular}{@{}l|rrr|c|cc|cc@{}}
				\toprule
				\multirow{2}{*}{\textbf{\begin{tabular}[c]{@{}l@{}}\quad Swin-L\\ \quad (198M)\end{tabular}}} & \multicolumn{1}{c}{\multirow{2}{*}{\textbf{\begin{tabular}[c]{@{}c@{}}Trained$\ast$ \\ Params \end{tabular}}}} & \multicolumn{1}{c}{\multirow{2}{*}{\textbf{\%}}} & 
				\multicolumn{1}{c|}{\multirow{2}{*}{\textbf{$\bm{\Delta_{Full}}$}}} &
				\multicolumn{1}{c|}{\multirow{2}{*}{\textbf{\begin{tabular}[c]{@{}c@{}}Extra\\ Structure\end{tabular}}}} & \multicolumn{2}{c|}{\textbf{\begin{tabular}[c]{@{}c@{}}Pascal VOC\\ (RetinaNet)\end{tabular}}} & \multicolumn{2}{c}{\textbf{\begin{tabular}[c]{@{}c@{}}ADE20K\\ (UperNet)\end{tabular}}} \\ \cline{6-9} 
				
				& \multicolumn{1}{c}{} & \multicolumn{1}{c}{} & \multicolumn{1}{c|}{} &  & \multicolumn{1}{c}{$\bm{\mathrm{AP_{Box}}}$} &\textbf{$\bm{\Delta_{Full}}$} & \multicolumn{1}{c}{$\bm{\mathrm{mIoU}}$} & \textbf{$\bm{\Delta_{Full}}$} \\ \midrule
				
				\rowcolor{gray!40}\multicolumn{9}{c}{\textit{\textbf{Baselines}}}\\ \midrule
				
				\quad {\scshape Full} & \multicolumn{1}{r}{ 198.58 M} & \multicolumn{1}{c}{100.00 \%} &\multicolumn{1}{c|}{-} & \multicolumn{1}{c|}{$\times$} & \multicolumn{1}{c}{83.70 \%} & \multicolumn{1}{c|}{-} & \multicolumn{1}{c}{ 51.18 \%}  & -\\
				\quad {\scshape Fixed} & 0.00 M & 0.00 \% &- 100.00 \% & $\times$ & \multicolumn{1}{c}{ 83.80 \%} & + 0.10 \%& 46.84 \% & - 4.34 \% \\ 
				\quad {\scshape BitFit} & 0.30 M & 0.15 \% & - 99.85 \%  & $\times$ & \multicolumn{1}{c}{85.40 \%} & + 1.70 \% & 48.37 \% & - 2.81 \% \\
				\quad {\scshape NormTuning} & 0.10 M & 0.05 \% & - 99.95 \% & $\times$ & \multicolumn{1}{c}{85.50 \%} & + 1.80 \% & 47.89 \%& - 3.29 \% \\ 
				\quad {\scshape Partial-1} & 28.77 M & 14.53 \% & - 85.47 \% & $\times$ & \multicolumn{1}{c}{85.50 \%} & + 1.80 \% & 47.44 \%& - 3.74 \% \\ \midrule
				\quad {\scshape Adapter} & 4.61 M & 2.33 \% & - 97.67 \% & $\checkmark$ & \multicolumn{1}{c}{86.70 \%} & + 3.00 \% & 50.78 \%& - 0.40 \% \\ 
				\quad {\scshape LoRA} & 4.57 M & 2.31 \% & - 97.69 \% & $\checkmark$ & \multicolumn{1}{c}{85.40 \%} & + 1.70 \% & 50.34 \%& - 0.84 \% \\ 
				\quad {\scshape AdaptFormer} & 2.34 M & 1.18 \% & - 98.82 \% & $\checkmark$ & \multicolumn{1}{c}{86.60 \%} & + 2.90 \% & 50.83 \%& - 0.35 \% \\ 
				\quad {\scshape LoRand} & 5.20 M & 2.62 \% & - 97.38 \% & $\checkmark$ & \multicolumn{1}{c}{86.90 \%} & + 3.20 \% & 50.93 \%& - 0.25 \% \\ \midrule

				\rowcolor{gray!40}\multicolumn{9}{c}{\textit{\textbf{Our Method}}}\\ \midrule
				
				\quad {\scshape \textbf{CoLin}} & 2.39 M & 1.23 \% & -98.77 \% & $\checkmark$ & \textbf{87.50 \%} & \textbf{ + 3.80 \%}& \textbf{51.28 \%} & \textbf{+0.10 \%} \\ \bottomrule
	\end{tabular}}}
	
    \label{tab:ade}
	
\end{table*}

\begin{table*}[tb]

  \centering
  \caption{\textbf{Results of baselines and our method on COCO benchmarks.} Swin-B is employed as the pre-trained model here. We present the numbers and percentages of trainable backbone parameters on the left and all the performences on the right. $\ast$ denotes the trainable parameters in backbones. The best AP in each column is bolded.}
  \scalebox{.8}{
		\setlength{\tabcolsep}{4.5mm}{
        
			\begin{tabular}{@{}l|rrr|c|cccc@{}}
				\toprule
				\multirow{2}{*}{\textbf{\begin{tabular}[c]{@{}l@{}}\quad Swin-B\\ \quad  (89M)\end{tabular}}} & \multicolumn{1}{c}{\multirow{2}{*}{\textbf{\begin{tabular}[c]{@{}c@{}}Trained$\ast$ \\ Params\end{tabular}}}} & \multicolumn{1}{c}{\multirow{2}{*}{\textbf{\%}}} & 
				\multicolumn{1}{c|}{\multirow{2}{*}{\textbf{$\bm{\Delta_{Full}}$}}} &
				\multicolumn{1}{c|}{\multirow{2}{*}{\textbf{\begin{tabular}[c]{@{}c@{}}Extra\\ Structure\end{tabular}}}} & \multicolumn{4}{c}{\textbf{\begin{tabular}[c]{@{}c@{}}COCO\\ (Cascade Mask R-CNN)\end{tabular}}}  \\ \cline{6-9} 
				
				& \multicolumn{1}{c}{} & \multicolumn{1}{c}{} & \multicolumn{1}{c|}{} &  & \multicolumn{1}{c}{$\bm{\mathrm{AP_{Box}}}$} &\textbf{$\bm{\Delta_{Full}}$} & \multicolumn{1}{c}{$\bm{\mathrm{AP_{Mask}}}$} & \textbf{$\bm{\Delta_{Full}}$} \\ \midrule
				
				\rowcolor{gray!40}\multicolumn{9}{c}{\textit{\textbf{Baselines}}}\\ \midrule
				\quad {\scshape Full} & \multicolumn{1}{r}{ 89.14 M} & \multicolumn{1}{c}{100.00 \%} &\multicolumn{1}{c|}{-} & \multicolumn{1}{c|}{$\times$} & \multicolumn{1}{c}{ 52.40 \%} & \multicolumn{1}{c}{-} & \multicolumn{1}{c}{ 45.10 \%}  & -\\
				\quad {\scshape Fixed} & 0.00 M & 0.00 \% &- 100.00 \% & $\times$ & \multicolumn{1}{c}{ 48.00 \%} & - 4.40 \%& 41.60 \% & - 3.50 \% \\ 
				\quad {\scshape BitFit} & 0.21 M & 0.23 \% & - 99.77 \%  & $\times$ & \multicolumn{1}{c}{50.10 \%} & - 2.30 \% & 43.60 \% & - 1.50 \% \\
				\quad {\scshape NormTuning} & 0.06 M & 0.07 \% & - 99.93 \% & $\times$ & \multicolumn{1}{c}{50.10 \%} & - 2.30 \% & 43.50 \%& - 1.60 \% \\ 
				\quad {\scshape Partial-1} & 12.95 M & 14.53 \% & - 85.47 \% & $\times$ & \multicolumn{1}{c}{50.60 \%} & - 1.80 \% & 43.70 \%& - 1.40 \% \\ \midrule
				\quad {\scshape Adapter} & 3.19 M & 3.58 \% & - 96.42 \% & $\checkmark$ & \multicolumn{1}{c}{52.10 \%} & - 0.30 \% & 45.00 \%& - 0.10 \% \\ 
				\quad {\scshape LoRA} & 3.06 M & 3.43 \% & - 96.57 \% & $\checkmark$ & \multicolumn{1}{c}{50.40 \%} & - 2.00 \% & 43.90 \%& - 1.20 \% \\ 
				\quad {\scshape AdaptFormer} & 1.60 M & 1.79 \% & - 98.21 \% & $\checkmark$ & \multicolumn{1}{c}{51.70 \%} & - 0.70 \% & 44.60 \%& - 0.50 \% \\ 
				\quad {\scshape LoRand} & 4.68 M & 5.23 \% & - 94.77 \% & $\checkmark$ & \multicolumn{1}{c}{51.90 \%} & - 0.50 \% & 44.70 \%& - 0.40 \% \\ \midrule

				\rowcolor{gray!40}\multicolumn{9}{c}{\textit{\textbf{Our Method}}}\\ \midrule
				
				\quad {\scshape \textbf{CoLin}} & 1.71 M & 1.97 \% &  -98.03 \% & $\checkmark$ & \textbf{52.90 \%} & \textbf{+ 0.50 \%}& \textbf{45.50 \%} & \textbf{ + 0.40 \%} \\ \bottomrule
				
		\end{tabular}
  }}
  
  \label{tab:coco}

\end{table*}

\section{Experiments}

\subsection{Datasets}
\label{dataset}
\noindent \textbf{Object Detection.} Pascal VOC 0712 \cite{everingham2015pascal} has 16k/5k training/validation images and is used for object detection tasks. We employ Swin-Large + RetinaNet for training. The evaluation metric for object detection task is the most commonly used AP$_{box}$.

\noindent \textbf{Semantic Segmentation.} ADE20K \cite{zhou2017scene} is a widely used semantic segmentation dataset containing 20K training and 2K validation images. We employ Swin-Large + UperNet for experiments on semantic segmentation. The evaluation metric is the most commonly used mIoU.


\noindent \textbf{Instance Segmentation.} MS COCO \cite{lin2014microsoft} is a representative instance segmentation dataset with 118k training images and 5k validation images. We employ Swin-Base + Cascade Mask RCNN for training. Evaluation metrics for instance segmentation task are AP$_{box}$ and AP$_{Mask}$.

\noindent \textbf{Oriented Object Detection.} Oriented object detection considers angle information in the annotation and inference process, which is more challenging than horizontal object detection. Two representative remote sensing datasets, DOTA \cite{dota} and STAR \cite{li2024star}, are selected for our experiments. We also experiment with multiple detection frameworks on the more challenging STAR dataset. The metric here is AP$_{box}$.

\noindent \textbf{Image Classification.} Classification tasks have been well studied in previous art. We also conduct experiments on Oxford 102 Flower \cite{nilsback2008automated}, Oxford-IIIT Pet  \cite{parkhi2012cats}, and VOC 2007 Classification dataset \cite{pascal-voc-2007} to increase the broadness of our experiments. The top-1, top-5, and average accuracy of each method are reported.

\begin{table*}[tb]

	\centering
    \caption{\textbf{Results of baselines and our method on three classification datasets.} Swin-L is employed as the pre-trained model here. We present top-1 accuracy (\%) and top-5 accuracy (\%) of each dataset. The best result in each column is bolded.}
	\scalebox{.8}{
		\setlength{\tabcolsep}{4.5mm}{
			\begin{tabular}{@{}lcccccccc@{}}
				\toprule
				\multirow{2}{*}{\quad \textbf{Method}} & \multicolumn{2}{c}{\textbf{Flowers102}}  & \multicolumn{2}{c}{\textbf{OxfordPets}}  & \multicolumn{2}{c}{\textbf{VOC2007} } & \multicolumn{2}{c}{\textbf{Average}} \\ \cmidrule(l){2-9} 
				& top-1 acc. & top-5 acc. & top-1 acc. & top-5 acc. & top-1 acc.     & top-5 acc. & top-1 acc.     & top-5 acc.    \\ \midrule
				
				\rowcolor{gray!30}\multicolumn{9}{c}{\textit{\textbf{Baselines}}}\\ \midrule
				
				\quad {\scshape Full}         &  99.5772  &  99.8536  &  94.6579  &  99.6257   &  84.1276  &  96.9507  & 92.7876 & 98.8100 \\
				\quad {\scshape Fixed}        &  99.3007  &  99.8374  &  94.2219  &  99.9182   &  85.0162  &  98.9499  & 92.8463 & 99.5685 \\
				\quad {\scshape BitFit}       &  99.5772  &  99.8211  &  95.3393  &  99.9182   &  85.6018  &  99.3336  & 93.5061 & 99.6910 \\
				\quad {\scshape NormTuning}   &  99.5284  &  99.8374  &  95.2303  &  99.8910   &  85.5210  &  99.2528  & 93.4266 & 99.6604 \\
				\quad {\scshape Partial-1}    &  99.6585  &  99.8374  &  95.3938  &  99.8637   &  84.9354  &  98.6066  & 93.3292 & 99.4359 \\
				\quad {\scshape Adapter}      &  99.5934  &  99.8536  &  95.3393  &  99.8092   &  87.0355  &  99.1317  & 93.9894 & 99.6144 \\
				\quad {\scshape LoRA}         &  99.5446  &  99.8536  &  95.1485  &  99.8910   &  85.7028  &  99.3134  & 93.4653 & 99.6860 \\
				\quad {\scshape AdaptFormer}  &  99.5609  &  99.8536  &  95.2576  &  99.8365   &  86.2884  &  99.2730  & 93.7023 & 99.6544 \\
				\quad {\scshape LoRand}       &  99.5725  &  99.8536  &  95.3515  &  99.8910   &  86.6534  &  99.3741  & 93.8591 & 99.7062 \\ \midrule
				\rowcolor{gray!30}\multicolumn{9}{c}{\textit{\textbf{Our Method}}}\\ \midrule
				\quad {\scshape \textbf{CoLin}}&  \textbf{99.6619}  &  \textbf{99.8816}  &  \textbf{95.4336}  & \textbf{99.9182}  &  \textbf{87.1132}
				&  \textbf{99.6108} & \textbf{94.0696}&\textbf{99.8035}\\ \bottomrule
				
	\end{tabular}}}
        
        \label{tab:cls}
\end{table*}

\begin{table}[tb]

  \centering
  \caption{\textbf{Results on DOTA and STAR benchmarks.} Swin-B is employed as the pre-trained model here. The best AP in each column is bolded.}
  \scalebox{.8}{
		\setlength{\tabcolsep}{0.9mm}{
			\begin{tabular}{@{}l|cc|c|c@{}}
				\toprule
				\multirow{2}{*}{\textbf{\begin{tabular}[c]{@{}l@{}}\quad Swin-B\\ \quad  (89M)\end{tabular}}} & \multicolumn{2}{c|}{\textbf{\begin{tabular}[c]{@{}c@{}}Oriented R-CNN\\(Faster R-CNN)\end{tabular}}} & \multicolumn{1}{c|}{\textbf{\begin{tabular}[c]{@{}c@{}}KLD\\(RetinaNet)\end{tabular}}} & \multicolumn{1}{c}{\textbf{\begin{tabular}[c]{@{}c@{}}H2RBox-v2 \\(FCOS)\end{tabular}}}   \\ \cline{2-5} 
				
				& \multicolumn{1}{c}{\textbf{DOTA-v1.0}} & \textbf{STAR} &   \textbf{STAR} &  \textbf{STAR}  \\ \midrule
				
				\rowcolor{gray!40}\multicolumn{5}{c}{\textit{\textbf{Baselines}}}\\ \midrule

                \quad {\scshape Full}                                                                             & 78.31  \%                                   & 38.63 \%                               & 30.33        \%                                                      & 30.29   \%                                                            \\
\quad {\scshape Fixed}                                                                            & 74.10            \%                         & 30.83             \%                   & 23.81    \%                                                          & 26.01       \%                                                        \\
\quad {\scshape BitFit}                                                                           & 76.05      \%                               & 34.51           \%                     & 28.17         \%                                                     & 29.41  \%                                                             \\
\quad {\scshape NormTuning}                                                                      & 75.82                 \%                    & 33.13         \%                        & 27.12            \%                                                   & 27.79            \%                                                    \\
\quad {\scshape Partial-1}                                                                       & 75.72            \%                          & 33.96           \%                      & 28.53                 \%                                              & 28.89    \%                                                            \\ \midrule
\quad {\scshape Adapter}                                                                         & 78.27         \%                             & 37.97                  \%               & 30.35                    \%                                           & 30.24     \%                                                           \\
\quad {\scshape LoRA}                                                                             & 75.91     \%                                 & 33.80               \%                  & 27.48         \%                                                      & 28.95     \%                                                           \\
\quad {\scshape AdaptFormer}                                                                      & 77.43     \%                                 & 35.95              \%                   & 29.36   \%                                                            & 30.11         \%                                                       \\
\quad {\scshape LoRand}                                                                           & 77.65        \%                              & 36.44           \%                      & 29.83       \%                                                        & 28.85  \%                                                              \\ \midrule
				
				\rowcolor{gray!40}\multicolumn{5}{c}{\textit{\textbf{Our Method}}}\\ \midrule
				
				\quad {\scshape \textbf{CoLin} }    & \textbf{ 78.39 \% }  & \textbf{39.22 \% }   & \textbf{30.68 \% }    & \textbf{ 31.14\% }               \\ \bottomrule
				
		\end{tabular}
  }}
  
    \label{tab:obb}
\end{table}

\subsection{Pre-trained Models and Toolkits}
\label{pretrained}

The Swin Transformer series \cite{liu2021swin} are employed as the backbone for all experiments. The pre-trained models are trained on ImageNet-22k \cite{deng2009imagenet}, and toolkits like MMDetection \cite{chen2019mmdetection}, MMSegmentation \cite{mmseg2020}, MMRotate \cite{zhou2022mmrotate} and MMClassification \cite{2020mmclassification} are used for verification. The image resolution of the pre-trained task is 224$\times$224. Most tasks employ Swin-Large as the backbone. Backbones for COCO, DOTA, and STAR are Swin-Base, considering the memory consumption of these tasks.

\subsection{Baselines}
\label{basline}

We compare CoLin with multiple recent methods. Baselines can be grouped into methods without or with extra structure.

\textbf{Without extra structure.} {\scshape Full}: Update all parameters in the framework. {\scshape Fixed}: Fix the backbone and update other parameters. {\scshape BitFit} \cite{zaken2021bitfit}: Update bias in backbone and parameters outside of backbone. {\scshape NormTuning} \cite{giannou2023expressive}: Update norm layers in backbone and parameters outside the backbone. {\scshape Partial-1} \cite{yosinski2014transferable}: Update the last block in the backbone and parameters outside the backbone.

\textbf{With extra structure.} 
{\scshape Adapter} \cite{houlsby2019parameter}: Add standard adapter layers after the MSA/MLP layers of each SwinBlock. {\scshape LoRA} \cite{hu2021lora}: Add parallel learnable matrices to multi-head attention weights. {\scshape AdaptFormer} \cite{chen2022adaptformer}: Add parallel adapter layers with scale weights to each MLP layer. {\scshape LoRand} \cite{yin20231}: Add LoRand++ ($\alpha$=4, $\beta$=16) layers after the MSA/MLP of each SwinBlock. LoRand++ has the best performance among its variants, so the most challenging setting is chosen for comparison. {\scshape Mona} \cite{yin20255}: Add Mona in the same way as the {\scshape Adapter} (Table \ref{tab:mona}).

\subsection{Main Results}
\label{results}
Table \ref{tab:ade} reports the performance of CoLin on object detection (Pascal VOC) and semantic segmentation (ADE20k) tasks. The results demonstrate that CoLin outperforms full fine-tuning and other representative methods while fine-tuning only 1.23\% of the backbone parameters. Notably, all PEFT methods surpass full fine-tuning on the relatively simple Pascal VOC dataset, serving as a low-resource setting in this context. This finding is consistent with results from the NLP domain (cite), corroborating that PEFT methods exhibit distinct advantages for few-shot learning tasks. Table \ref{tab:coco} extends the evaluation to CoLin’s performance on fine-grained instance segmentation using the COCO dataset. Similarly, CoLin achieves superior performance to full fine-tuning and competing approaches by fine-tuning only 1.97\% of new parameters. Given that instance segmentation imposes more stringent demands on model architectures than object detection and standard semantic segmentation, the results in Table \ref{tab:coco} more robustly demonstrate CoLin’s efficacy on complex vision tasks. As image classification tasks receive considerable attention in the research community, we present the performance of all methods across three widely used classification datasets in Table \ref{tab:cls}. The results indicate that CoLin also delivers promising performance in image classification. Notably, due to the narrow performance gaps among these methods on classification datasets, our focus is directed toward their performance differences on complex tasks. Lastly, Table \ref{tab:obb} evaluates the performance of all methods on the remote sensing rotated object detection task. Since pre-trained models are initialized on natural-scene data, these results also characterize the cross-domain transfer capabilities of the evaluated methods. Specifically, results in Table \ref{tab:obb} demonstrate that CoLin maintains its superiority over full fine-tuning and representative baselines in cross-domain transfer scenarios.

\begin{table}[tb]

	\centering
    \caption{\textbf{Comparisons with Mona.} CoLin outperforms Mona on param flexibility and inference speed (tested on 8$\times$ RTX 4090).}
	\scalebox{.8}{\setlength{\tabcolsep}{.9mm}{
	\begin{tabular}{@{}lcccc@{}}
\toprule
\textbf{Module}            & \textbf{Task/Backbone} & \textbf{\%\bm{$_{Full}$}} & \textbf{Per.} & \textbf{Inference Speed} \\ \midrule
{\scshape Mona }          & COCO/Swin-B   & 4.67\%   & 53.4\%      & 40.1 tasks/s\\
{\scshape CoLin} (D128K64) & COCO/Swin-B   & 4.59\%   & 53.3\%     & 46.3 tasks/s      \\
{\scshape CoLin} (D100K28) & COCO/Swin-B   & 1.97\%   & 52.9\%     & 47.5 tasks/s     \\ \midrule
{\scshape Mona }          & VOC/Swin-L    & 2.56\%   & 87.3\%      & 61.9 tasks/s      \\
{\scshape CoLin} (D128K64) & VOC/Swin-L    & 2.52\%   & 87.3\%     & 73.8 tasks/s      \\
{\scshape CoLin} (D100K28) & VOC/Swin-L    & 1.23\%   & 87.5\%     & 75.1 tasks/s     \\ \bottomrule
\end{tabular}}}
        
        \label{tab:mona}
	
 \vspace{-10pt}
\end{table}

Beyond the above methods, we conduct a multi-dimensional comparison between CoLin and Mona \cite{yin20255}. For a more intuitive comparison, Table \ref{tab:mona} presents the parameter count, performance, and inference speed of two variants of CoLin against Mona under different settings. When parameter are comparable, CoLin achieves performance on par with Mona. CoLin’s advantages over Mona are twofold: 1) CoLin can flexibly adjust the number of new parameters, thereby regulating the parameter redundancy of the adapter. Results on Pascal VOC demonstrate that fewer parameters yield superior performance in few-shot scenarios. 2) CoLin exhibits lower inference latency than Mona. Specifically, Mona incorporates multiple DWConv layers, one pointwise conv, and two linear projections, whereas CoLin only involves one DWConv layer and two linear projections during inference. Notably, the multi-branch structure of CoLin employed in training phase can be converted into standard linear layers during inference.  \textbf{Overall, CoLin boosts inference speed by nearly 20\% than Mona} while still achieving impressive performance, which significantly lowers the per-inference cost for commercial enterprises.

\begin{figure}[tb]
	\centering
	\includegraphics[scale=.27]{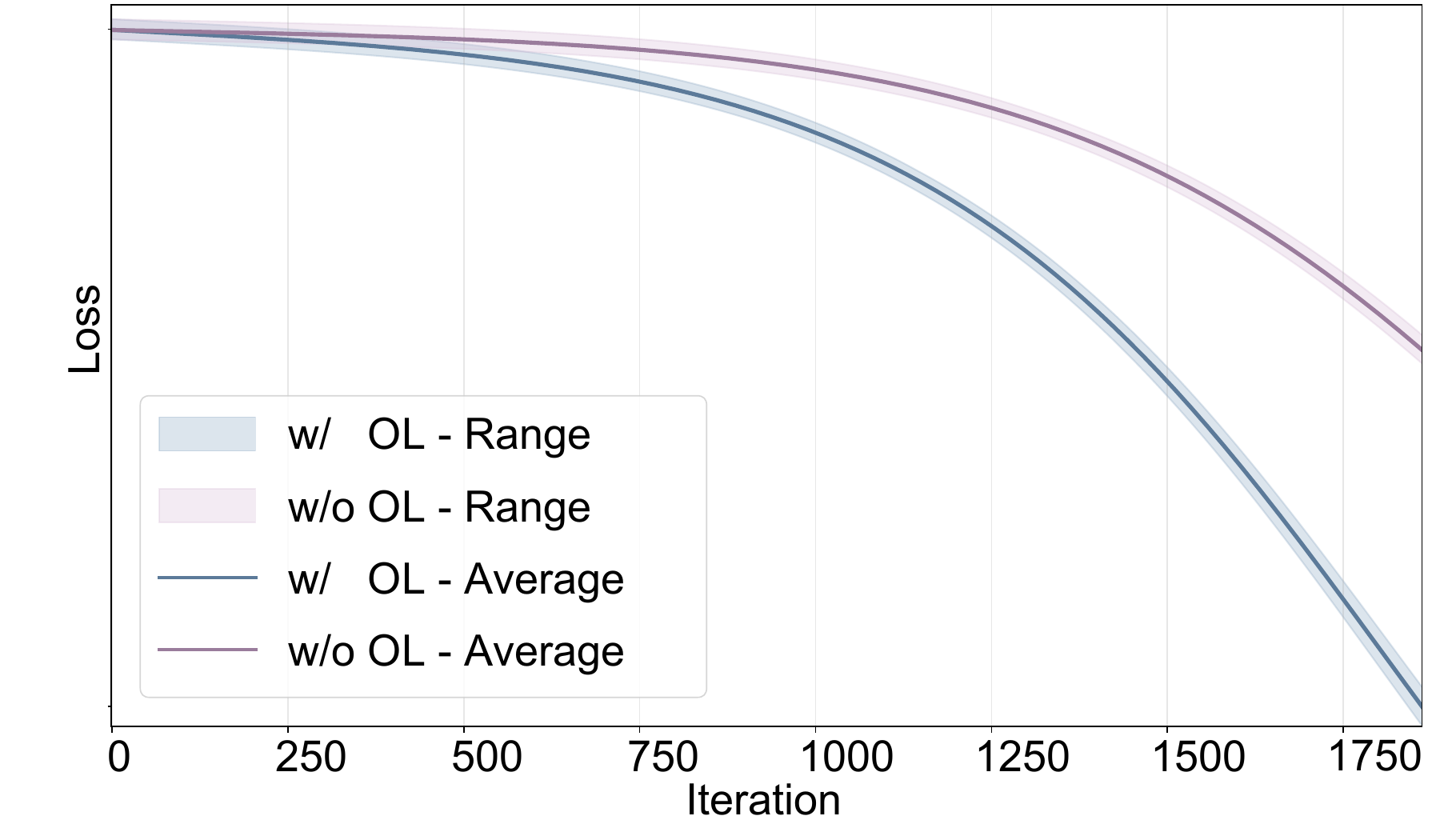}
	\caption{\textbf{Convergence simulation.} Theoretical analysis and simulation experiments demonstrate that in low-rank synthesis scenarios, orthogonality (blue) significantly enhances the overall convergence efficiency of the model.
}
    \vspace{-10pt}
	\label{fig:simulate}
\end{figure}

In addition to the comparative experiments, we design a simulation experiments to illustrate the role of orthogonal loss in the convergence process of low-rank composite matrices. Specifically, we devise a simple loss function to drive two groups of matrices to approximate a randomly generated target matrix \(W \in \mathbb{R}^{m \times n}\) via gradient descent. For matrices \(P \in \mathbb{R}^{k \times m}\) and \(Q \in \mathbb{R}^{k \times n}\), we minimize the loss function \(\text{Loss} = \|W - P^\top Q\|_2\) to make \(P^\top Q\) approach \(W\). We set the hyperparameters as \([m, k, n] = [100, 30, 5000]\) and the learning rate (lr) to \(10^{-5}\), then calculate the mean and range of the loss values over 20 random seeds. Figure \ref{fig:simulate} depicts the loss variations of the two matrix groups during convergence, where the blue curve corresponds to the setup with the additional orthogonal loss (OL) and the purple curve denotes the setup without OL. As can be observed, the convergence efficiency of the blue group is significantly higher than that of the purple group at the same number of iterations. Section \ref{PSQ} and Figure \ref{fig:simulate} clarify the convergence issues of low-rank composite matrices from both theoretical and experimental perspectives, which effectively explains the superior performance of CoLin. Appendix \ref{appen-sec:sim} provides simulation results on more matrix sizes. Additionally, we discuss limitations and future directions in Appendix \ref{sec:limitation}.

\subsection{Ablations}

\textbf{Internal Design.} Table \ref{tab:ab} presents the ablation results of CoLin during the optimization process. Vanilla version denotes the multi-branch low-rank adapter architecture with a sharing mechanism (as illustrated in Figure \ref{fig:detail}). OL refers to the orthogonal loss introduced in Section \ref{PSQ}, while SVDinit denotes the initialization method described in Section \ref{SVD}. Without matrix optimization, CoLin achieves performance comparable to full fine-tuning by tuning only about 1\% of the total parameters (see Tables \ref{tab:coco} and \ref{tab:ade}), but fails to outperform it. The incorporation of orthogonal loss enables CoLin to surpass full fine-tuning on both few-shot and complex tasks. The SVD initialization method facilitates the low-rank matrices to converge toward orthogonality, which further enhances the convergence efficiency of the orthogonal loss. Table \ref{tab:ab} demonstrate that the optimization designs proposed in this work can effectively boost the performance of visual adapters, which experimentally validates the theoretical analysis presented in Section \ref{PSQ}.

\begin{table}[tb]

	\centering
    \caption{\textbf{Design ablations.} Multiple necessary components enable CoLin to outperform full fine-tuning with 1\% new parameters.}
	\scalebox{.8}{\setlength{\tabcolsep}{6mm}{
	\begin{tabular}{@{}lcc@{}}
    \toprule
    \textbf{Method}         & \textbf{Pascal VOC}        & \textbf{COCO} \\ \midrule
    {\scshape Vallina}        & 87.0 \%       & 52.0 \%\\
    {\scshape +OL}            & 87.3 \%      & 52.6 \%\\
    {\scshape +OL+SVDinit}    & 87.5 \%      & 52.9 \%\\
    \bottomrule
    \end{tabular}}}
        
        \label{tab:ab}
	
\end{table}

\begin{table}[h]

	\centering
    \caption{\textbf{Ablations on hyperparameters.} Results under different value of $D$, $K$, and $B$.}
	\scalebox{.8}{\setlength{\tabcolsep}{4.6mm}{
	\begin{tabular}{@{}c|ccc@{}}
    \toprule
    \textbf{Param}    & \textbf{Value} & \textbf{\%\bm{$_{Full}$}} & \textbf{Per. On VOC} \\ \midrule
    \multirow{3}{*}{$D$} & 128   & 1.27\%        & 87.5\%      \\
                       & 100   & 1.23\%        & 87.5\%      \\
                       & 72    & 1.18\%        & 87.2\%      \\ \midrule
    \multirow{3}{*}{$K$} & 32    & 1.40\%        & 87.6\%      \\
                       & 28    & 1.23\%        & 87.5\%      \\
                       & 16    & 0.72\%        & 86.9\%      \\ \midrule
    \multirow{3}{*}{$B$} & 8     & 1.23\%        & 87.5\%      \\
                       & 4     & 1.23\%        & 87.5\%      \\
                       & 1     & 1.23\%        & 87.2\%      \\ \bottomrule
    \end{tabular}}}
        
        \label{tab:ab-hp}
	
\end{table}

\textbf{Hyperparameter.} We conduct ablation studies on three key hyperparameters. $D$ denotes the projection dimension, $K$ denotes the kernel size, and $B$ denotes the number of branches. For all experiments, the baseline configuration is set as $[D, K, B] = [100, 28, 4]$, where only one of the three hyperparameters is varied in each individual trial. Results in Table \ref{tab:ab-hp} show that kernel size ($K$) exerts the most significant impact on the number of trainable parameters. By contrast, adjustments to $D$ and $B$ yield negligible effects on the trainable parameter count, owing to the sharing mechanism. Considering the trade-off between parameter count and model performance, we ultimately adopt $[D, K, B] = [100, 28, 4]$ as the standard configuration for CoLin.

\section{Conclusion}

We propose CoLin, which surpasses full fine-tuning with only about 1\% backbone params on visual recognition tasks for the first time. With matrix theory, we analyze and optimize the gradient backpropagation process of the low-rank composite matrices, effectively enhancing CoLin’s transfer efficiency of visual features. Extensive experiments demonstrate that the proposed method outperforms full fine-tuning across a broad range of visual recognition tasks. As large vision models continue to expand in parameter and knowledge capacity, the transfer potential of CoLin is poised to be further amplified for both scholars and developers.

\section*{Impact Statement}

This paper presents work whose goal is to advance the field of Machine
Learning. There are many potential societal consequences of our work, none
which we feel must be specifically highlighted here.

\bibliography{example_paper}
\bibliographystyle{icml2026}

\begin{table*}[tb] 
\centering
\caption{\textbf{Definitions of core symbols.}}
\scalebox{1}{
		\setlength{\tabcolsep}{1mm}
        {
\begin{tabular}{ll}
\toprule
Symbol          & Description                                                                 \\
\midrule
$W$             & Projection matrix in the adapter (dimension $m \times n$), with the goal of minimizing the loss $L$ by optimizing $W$ \\
$P, Q$          & Low-rank decomposition factors: $P \in \mathbb{R}^{\beta \times m}$, $Q \in \mathbb{R}^{\beta \times n}$, satisfying $W = PQ$ \\
$L(W)$          & Task loss function (scalar), dependent on the projection matrix $W$     \\
$\nabla_W L$    & Gradient of the loss $L$ with respect to $W$ (dimension $m \times n$) \\
$\eta$          & Learning rate (positive number, typically in the range $10^{-5} \sim 10^{-3}$, satisfying $\eta^2 \ll \eta$) \\
$\text{tr}(\cdot)$ & Matrix trace (sum of diagonal elements)                                     \\
$\|\cdot\|_F$   & Frobenius norm (square root of the sum of squares of all matrix elements)   \\
$I_k$           & $k$-dimensional identity matrix (here $k = \beta$, i.e., the rank of the low-rank decomposition) \\
\bottomrule
\end{tabular}
}}
\end{table*}

\newpage
\appendix

\section{Detailed proof of Section \ref{PSQ}}
\label{proof}

\subsection*{A1. Assumptions and Notation Definitions}
To rigorously complete the proof, we first clarify the definitions of core symbols and basic assumptions in this section:

\paragraph{Basic Assumptions:}
1. The matrix differential satisfies the product rule: For any differentiable matrices $A, B$, $d(AB) = A dB + dA B$; \\
2. The learning rate $\eta$ is sufficiently small, and the higher-order infinitesimal term $\eta^2$ can be neglected; \\
3. The loss function $L(W)$ is twice differentiable with respect to $W$, and the gradient $\nabla_W L$ exists and is continuous.

\subsection*{A2. Update Rule of $W$ in the Ideal Case}
The core idea of gradient descent is to update parameters along the negative gradient direction of the loss function to minimize the loss. For the projection matrix $W$, the ideal update rule is:
\[ W' = W - \eta \nabla_W L \],
where $W'$ denotes the updated matrix. Defining the parameter update amount $\Delta W = W' - W$, we have in the ideal case:
\begin{equation} \tag{A1}
\Delta W = -\eta \nabla_W L.
\end{equation}
At this point, the differential of the loss is $dL = \text{tr}(\nabla_W^T \Delta W) = -\eta \text{tr}(\nabla_W^T \nabla_W)$. Since $\text{tr}(\nabla_W^T \nabla_W) = \|\nabla_W\|_F^2 \geq 0$, the loss can decrease with maximum efficiency.

\subsection*{A3. Derivation of the Actual $\Delta W$ Under Low-Rank Decomposition}
When $W = PQ$, the actual objects of parameter update are the low-rank factors $P$ and $Q$, rather than $W$ itself. Assume the update rules for $P$ and $Q$ are:
\begin{equation} \tag{A2}
P' = P - \eta \nabla_P L, \quad Q' = Q - \eta \nabla_Q L,
\end{equation}
where $\nabla_P L$ and $\nabla_Q L$ are the gradients of the loss $L$ with respect to $P$ and $Q$, respectively.

\subsubsection*{A3.1 Expansion of $\Delta W$}
The updated $W' = P'Q'$, thus:

\begin{align*}
\Delta W &= W' - W = (P - \eta \nabla_P L)(Q - \eta \nabla_Q L) - PQ \\
&= PQ - \eta \nabla_P L \cdot Q - \eta P \cdot \nabla_Q L + \eta^2 \nabla_P L \cdot \nabla_Q L - PQ \\
&= -\eta \left( \nabla_P L \cdot Q + P \cdot \nabla_Q L \right) + \eta^2 \nabla_P L \cdot \nabla_Q L.
\end{align*}

Since the learning rate $\eta \ll 1$, the contribution of the higher-order term $\eta^2 \nabla_P L \cdot \nabla_Q L$ to $\Delta W$ is negligible. Therefore:
\begin{equation} \tag{A3}
\Delta W \approx -\eta \left( \nabla_P L \cdot Q + P \cdot \nabla_Q L \right).
\end{equation}

\subsubsection*{A3.2 Derivation of Gradients $\nabla_P L$ and $\nabla_Q L$}
To solve $\nabla_P L$ and $\nabla_Q L$, we need to use the \textit{trace trick} for matrix derivatives, which relies on two key properties of the trace operation:\\
1. Cyclic permutation property: For any compatible matrices $A, B$, $\text{tr}(AB) = \text{tr}(BA)$; \\
2. Linearity: For any matrices $A, B$ and constant $c$, $\text{tr}(A+B) = \text{tr}(A) + \text{tr}(B)$ and $\text{tr}(cA) = c \text{tr}(A)$.

\paragraph{Step 1: Definition of Loss Differential}
For the loss function $L(W)$, its differential can be expressed as the inner product of the gradient and the parameter differential, in matrix form:
\begin{equation} \tag{A4}
dL = \text{tr}\left( \nabla_W^T dW \right),
\end{equation}
where $dW$ is the differential matrix of $W$.

\begin{figure*}[h]
	\centering
	\includegraphics[scale=.62]{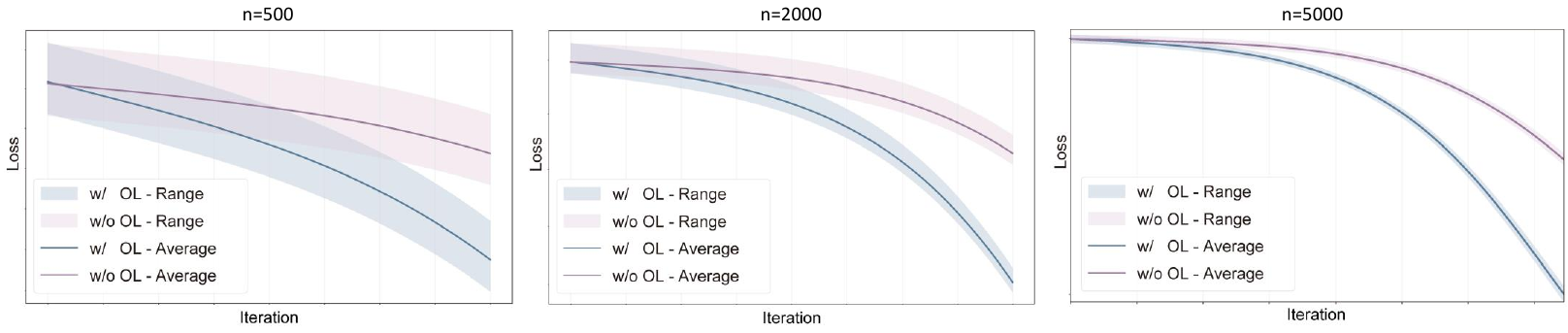}
	\caption{\textbf{Impact of Orthogonal Loss on Matrices of Different Sizes.} All iterations range from 0 to 2000. The difference among the three figures lies in the second dimension size of the matrices. As the size increases, the benefit of adding orthogonal loss becomes greater.}
	\label{fig:appen-sim}
\end{figure*}

\paragraph{Step 2: Substitute the Low-Rank Decomposition $W = PQ$}
According to the product rule of matrix differentials, $dW = P dQ + dP Q$ (where $dP$ and $dQ$ are the differential matrices of $P$ and $Q$, respectively). Substituting this into Equation (A4):
\[ dL = \text{tr}\left( \nabla_W^T (P dQ + dP Q) \right) = \text{tr}\left( \nabla_W^T P dQ \right) + \text{tr}\left( \nabla_W^T dP Q \right). \]

\paragraph{Step 3: Differentiate with Respect to $dQ$ (Derive $\nabla_Q L$)}
Applying the cyclic permutation property of the trace to the first term above:
\[ \text{tr}\left( \nabla_W^T P dQ \right) = \text{tr}\left( (P^T \nabla_W)^T dQ \right). \]
By the definition of matrix gradient: If $dL = \text{tr}(G^T dX)$, then $\nabla_X L = G$. Here, $X = Q$, so:
\begin{equation} \tag{A5}
\nabla_Q L = P^T \nabla_W.
\end{equation}

\paragraph{Step 4: Differentiate with Respect to $dP$ (Derive $\nabla_P L$)}
Applying the cyclic permutation property of the trace to the second term above:
\[ \text{tr}\left( \nabla_W^T dP Q \right) = \text{tr}\left( Q \nabla_W^T dP \right) = \text{tr}\left( (\nabla_W Q^T)^T dP \right). \]
Similarly, by the definition of matrix gradient, $X = P$, so:
\begin{equation} \tag{A6}
\nabla_P L = \nabla_W Q^T.
\end{equation}

\subsubsection*{A3.3 Simplification of $\Delta W$ After Substituting Gradients}
Substituting Equations (A5) and (A6) into Equation (A3):

\begin{align*} \tag{A7}
\Delta W &\approx -\eta \left( (\nabla_W Q^T) \cdot Q + P \cdot (P^T \nabla_W) \right) \\
&= -\eta \left( \nabla_W (Q Q^T) + (P^T P) \nabla_W \right).
\end{align*}

Due to the associativity of matrix multiplication, $\nabla_W (Q Q^T) = \nabla_W \cdot Q Q^T$ (where $Q Q^T$ is a $\beta \times \beta$ matrix, compatible with the left multiplication of $\nabla_W$), and $(P^T P) \nabla_W = P^T P \cdot \nabla_W$. Thus, the common factor $\nabla_W$ can be factored out:
\begin{equation} \tag{A8}
\Delta W \approx -\eta \nabla_W \left( Q Q^T + P^T P \right).
\end{equation}

\subsection*{A4. Impact of Orthogonality on Optimization Efficiency}
\subsubsection*{A4.1 Core Criterion for Optimization Efficiency}
The efficiency of loss reduction is determined by $dL = \text{tr}(\nabla_W^T \Delta W)$: When $\Delta W$ is aligned with $-\nabla_W$, $dL$ reaches its minimum (the loss decreases the fastest). Combining Equation (A8):
\[ dL \approx -\eta \text{tr}\left( \nabla_W^T \nabla_W (Q Q^T + P^T P) \right). \]
If $Q Q^T + P^T P = c I_\beta$ (where $c > 0$ is a constant), then:
\[ dL \approx -\eta c \text{tr}\left( \nabla_W^T \nabla_W \right). \]
At this point, $\Delta W$ is completely aligned with $-\nabla_W$, and the optimization efficiency is maximized (the constant $c$ only affects the step size, which can be adjusted via the learning rate $\eta$).

\subsubsection*{A4.2 Optimality of Orthogonality}
When $P$ and $Q$ satisfy the \textit{orthogonality condition}:
\begin{equation} \tag{A10}
P^T P \approx I_\beta, \quad Q Q^T \approx I_\beta,
\end{equation}
we have $Q Q^T + P^T P \approx I_\beta + I_\beta = 2 I_\beta$. Substituting into Equation (A8):
\begin{equation} \tag{A11}
\Delta W \approx -2\eta \nabla_W.
\end{equation}
At this point, $\Delta W$ is completely consistent with the ideal update direction (Equation (A1)) (only the step size coefficient is 2, which does not affect the direction), and the loss reduction efficiency is maximized.

Conversely, if $P$ or $Q$ is non-orthogonal (e.g., $P^T P = 3I_\beta$, $Q Q^T = 0.5I_\beta$), then $Q Q^T + P^T P = 3.5I_\beta$, leading to $\Delta W \approx -3.5\eta \nabla_W$, where the step size deviates from the optimal value. More critically, if $Q Q^T + P^T P$ is a non-diagonal matrix, the direction of $\Delta W$ will deviate from $-\nabla_W$, significantly reducing optimization efficiency and even causing training oscillations.

\subsection*{A6. Conclusion}
The detailed proof above demonstrates that the low-rank decomposition $W = PQ$ causes the parameter update direction to deviate from the ideal negative gradient direction. The orthogonality of $P$ and $Q$ is the key to solving this problem—orthogonality enables $\Delta W$ to approximate the ideal update direction, maximizing optimization efficiency. By introducing the orthogonal regularization loss, this orthogonality can be maintained during training, ultimately improving the convergence speed and performance of the low-rank adapter.

\section{The Impact of Orthogonality on Matrices of Different Sizes}
\label{appen-sec:sim}

Figure \ref{fig:simulate} illustrates the convergence processes under the setting of $[m, k, n] = [100, 30, 5000]$. To further investigate the impact of orthogonal loss on matrices of different sizes, we conduct experiments across a broader range of size configurations. From left to right, Figure \ref{fig:appen-sim} presents the convergence processes corresponding to matrix sizes $n=[500, 2000, 5000]$. All three sets of results are obtained from experiments with 20 random seeds. It can be observed that orthogonality yields a more pronounced improvement in convergence for larger matrices. Low-rank synthesis is prevalent in the era of large models (e.g., LoRA, LoRand, CoLin). As the dimensionality of large models continues to increase, orthogonal loss will bring greater benefits to model convergence.

\section{Limitation}
\label{sec:limitation}
We aim to discuss the limitations herein to identify promising avenues for future exploration in this domain. To the best of our knowledge, there is no method has been shown to outperform full fine-tuning across a wide range of visual recognition tasks with extra latency. While adapter-based approaches yield impressive performance on visual tasks, they inevitably introduce extra modules and inference latency. We observe that LoRA-like methods often outperform full fine-tuning in large language models (LLMs) without incurring additional inference costs \cite{ding2023parameter}. This is primarily because the input and output distributions of LLMs are nearly homogeneous. LoRA excels at LLMs due to its ability to enable efficient domain transfer for homogeneous data. For visual recognition tasks, however, the input and output distributions of foundation models during pretraining (e.g., classification, unsupervised learning) differ significantly from those of downstream tasks (e.g., detection, segmentation). Thus, fine-tuning visual foundation models for downstream tasks imposes more stringent requirements on tuning methods than fine-tuning LLMs does. Going forward, research on fine-tuning for visual recognition tasks should strive toward the goal of zero inference latency—a direction that we also intend to pursue in our future work.

\begin{figure}[tb]
	\centering
	\includegraphics[scale=.7]{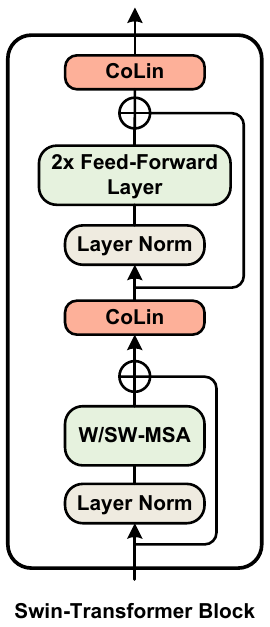}
	\caption{\textbf{Insertion location.} CoLin is inserted after the two skip connections in each SwinBlock.}
	\label{fig:swin}
\end{figure}

\section{Adapter Tuning Paradigm}
\label{appen-paradigm}
For dataset $D=\left\{(x_i,y_i)\right\}^N_{i=1}$, fine-tuning calculates the loss between inference results and labels according to the formula:
\[ L\left(D,\theta\right)=\sum_{i=1}^{N}{loss(f_\theta\left(x_i\right),y_i)}, \]

where $f_\theta$ denotes the network forward function and $loss$ represents the loss function. After that, $\theta$ is optimized through 

\[ \theta\gets\underset{\theta}{{\arg\min} \, L(D,\theta)}.\]

In adapter tuning paradigm, parameters consist of two parts, including parameters in adapter $\theta_A$ and parameters in the original architecture $\theta$. Here, $\theta$ is further divided into frozen part $\theta_F$ and trainable part $\theta_T$, noted as $\theta=\left\{\theta_F, \theta_T \right\}$. Let $\Omega$ be all the trainable parameters, then $\Omega=\left\{\theta_A, \theta_T \right\}$. The loss function and optimization formula in adapter can be written as:
\[ L\left(D,\theta_F,\Omega\right)=\sum_{i=1}^{N}{loss(f_{\theta_F,\Omega}\left(x_i\right),y_i)},\]

\[ \Omega\gets\underset{\Omega}{{\arg\min} \, L(D,\theta_F, \Omega).}\]

\section{The Position of CoLin in Swin}
\label{pos}

Our experiments are conducted based on the Swin Transformer (Swin-B and Swin-L). We insert CoLin sequentially after the two skip connections following the configuration illustrated in Figure \ref{fig:swin}.



\end{document}